\documentclass[12pt,leqno]{article}
\usepackage{latexsym}
\newtheorem{theorem}{Theorem}
\newtheorem{corollary}{Corollary}
\newtheorem{lemma}{Lemma}

\newtheorem{definition}{Definition}
\newtheorem{example}{Example}

\newenvironment{notation}{\noindent\bf Notation:\em\penalty100}{}
\newcommand{\blackslug}{\mbox{\hskip 1pt \vrule width 4pt height 8pt 
depth 1.5pt \hskip 1pt}}
\newcommand{\QED}{\quad\blackslug\lower 8.5pt\null\par\noindent}
\newcommand{\proof}{\par\penalty-100\vskip .5 pt\noindent{\bf Proof\/: }}
\newcommand{\cC}{\mbox{${\cal C}$}}
\newcommand{\cH}{\mbox{${\cal H}$}}
\newcommand{\cL}{\mbox{${\cal L}$}}
\newcommand{\cM}{\mbox{${\cal M}$}}
\newcommand{\Cn}{\mbox{${\cal C}n$}}
\newcommand{\subseteqf}{\mbox{$\subseteq_{f}$}}
\newcommand{\eqdef}{\stackrel{\rm def}{=}}
\newcommand{\ru}{\rule[-0.4mm]{.1mm}{3mm}}
\newcommand{\nni}{\ru\hspace{-3.5pt}}

\newcommand{\pre}{\hspace{0.28em}}

\newcommand{\NIm}{\pre\nni\sim}
\newcommand{\NI}{\mbox{$\: \nni\sim$}}

\title{Connectives in Quantum and other Cumulative Logics
\thanks{This work was partially supported 
by the Jean and Helene Alfassa fund for 
research in Artificial Intelligence.}
}
\author{Daniel Lehmann\\
School of Computer Science and Engineering, \\Hebrew University, \\Jerusalem 91904, Israel
\\lehmann@cs.huji.ac.il
}
\date{May 2002}

\begin{document}
\maketitle
\begin{abstract}
The nonmonotonic logics definable by definability-preserving 
choice functions that satisfy Coherence 
have been studied in~\cite{L:LogicsandSemantics}. 
Larger families correspond to weakenings of this property. 
The cumulative and loop-cumulative relations of~\cite{KLMAI:89} 
are characterized by such models and, as a consequence, 
one may study the natural connectives for those logics. 
The representation results obtained are surprisingly smooth: 
in the completeness part the choice function may be defined on any set of models, 
not only definable sets and no definability-preservation property is required
in the soundness part.
For those logics, proper conjunction and negation may be defined, 
but no proper disjunction, 
contrary to the situation studied in~\cite{L:LogicsandSemantics}. 
Quantum Logics, as defined by~\cite{EngGabbay:Quantum} are such Logics but 
the orthogonal complement does not provide a proper negation.
\end{abstract}
\section{Introduction}
In~\cite{BirkvonNeu:36}, Birkhoff and von Neumann suggested that 
the logic of quantum mechanics be isomorphic to 
the algebra of closed subspaces of Hilbert spaces, under ``set product'' 
(i.e., intersection), ``closed linear sum'', and ``orthogonal
complement''. Many researchers studied the properties of those operations 
and their results are reviewed in~\cite{DallaChiara:01}.
Recently, Engesser and Gabbay~\cite{EngGabbay:Quantum} proposed a very different and deeper connection between Logics and Quantum mechanics. 
For them every quantum state defines a consequence relation.
They showed that those consequence relations are 
nonmonotonic and enjoy some of the most important properties studied 
in~\cite{KLMAI:89}, in particular cumulativity.
Whereas Engesser and Gabbay assume a language closed under the 
propositional connectives (as did Birkhoff and von Neumann), even though those
connectives are not at all classical, 
the purpose of this paper is to study and try to characterize 
the consequence operations presented by Quantum mechanics
before any connectives
are defined, in the style of the author's~\cite{L:LogicsandSemantics}.
Since Quantum Logics fail, in general, to satisfy two of the properties assumed
there, representation results for larger families than those of~\cite{L:LogicsandSemantics} are needed. Such results will be developed first. 
For the conservative extension results to be proven below, 
models closer to the cumulative models of~\cite{KLMAI:89} 
or of~\cite{Mak:Handbook} could have been used. 
The models presented here and their tight link with the failure of Coherence 
have been preferred both for their intrinsic interest and for compatibility 
with~\cite{L:LogicsandSemantics}.
\subsection{Reflections on this paper}
The study of C-logics is unexpectedly smooth and attractive.
The basic intuition behind the cumulative relations of KLM is confirmed:
cumulative relations yield classical connectives but the disjunction 
(that may be defined as usual from negation and conjunction) does not
behave proof-theoretically as a proper disjunction should.
The section on L-logics is less interesting. I am not sure where it leads.
The results are straightforward translation from KLM and L-logics do not seem
to behave in any better way with respect to connectives than C-logics.
The reason it may be interesting is that Quantum Logics are not only C-logics but
also L-logics.
But two main questions are left open: can all L-logics be presented as Quantum
Logics or do Quantum Logics satisfy additional properties? 
What is the meaning for Quantum Logics of the classical negation and conjunction
that can be defined for any C-logics? Why do Birkhoff and von Neumann expect 
that the negation of an observable be observable?
\section{C-logics}
\subsection{Definition}
The framework is the one presented in~\cite{L:LogicsandSemantics}.
Let \cL\ be any non-empty set. The elements of \cL\ should be viewed as
propositions or formulas and \cL\ is therefore a language. At present no 
structure is assumed on \cL\ and its elements are therefore to be taken as 
atomic propositions.
Let \mbox{$\cC : 2^{\cL} \longrightarrow 2^{\cL}$}.
\begin{definition}
\label{def:C-logics}
The operation \cC\ is said to be a {\em C-logics} iff it satisfies the 
two following properties.
\[
{\bf Inclusion} \ \ \ \forall A \subseteq \cL\ , \ 
A \subseteq \cC(A) , 
\]
\[
{\bf Cumulativity} \ \ \ \forall A , B \subseteq \cL , \ \  
A \subseteq B \subseteq \cC(A) 
\Rightarrow \cC(A) = \cC(B).
\]
\end{definition}
\subsection{Properties}
\begin{lemma}[Makinson]
\label{le:idem}
An operation \cC\ is a C-logics iff it satisfies Inclusion,
\[
{\bf Idempotence} \ \ \ \forall A \subseteq \cL , \ \ 
\cC(\cC(A)) = \cC(A) 
\]
and
\[
{\bf Cautious \ Monotonicity} \ \ \ \forall A , B \subseteq \cL\ 
A \subseteq B \subseteq \cC(A) 
\Rightarrow \cC(A) \subseteq \cC(B) 
\]
\end{lemma}
\proof
Let us prove, first, that a C-logics satisfies Idempotence.
By Inclusion \mbox{$A \subseteq \cC(A) \subseteq \cC(A)$}, therefore,
by Cumulativity: \mbox{$\cC(A) = \cC(\cC(A))$}.
Assume, now that \cC\ satisfies Inclusion, Idempotence and 
Cautious Monotonicity. Let \mbox{$A \subseteq B \subseteq \cC(A)$}.
By Cautious Monotonicity, we have \mbox{$\cC(A) \subseteq \cC(B)$}.
Therefore, we have \mbox{$B \subseteq \cC(A) \subseteq \cC(B)$}.
By Cautious Monotonicity again, we have: 
\mbox{$\cC(B) \subseteq \cC(\cC(A))$}. By Idempotence, then, we conclude
\mbox{$\cC(B) \subseteq \cC(A)$} and therefore 
\mbox{$\cC(B) = \cC(A)$}.
\QED
\begin{lemma}[Makinson]
\label{le:2-loop}
An operation \cC\ is a C-logics iff it satisfies Inclusion and
\[
{\bf 2-Loop} \ \ \ \ A \subseteq \cC(B) , B \subseteq \cC(A) \Rightarrow 
\cC(A) = \cC(B).
\]
\end{lemma}
\proof
Assume \cC\ is a C-logics and \mbox{$A \subseteq \cC(B)$}. By Inclusion,
we have \mbox{$B \subseteq A \cup B \subseteq \cC(B)$} and by Cumulativity:
\mbox{$\cC(B) = \cC(A \cup B)$}.
Similarly \mbox{$B \subseteq \cC(A)$} implies 
\mbox{$\cC(A) = \cC(A \cup B)$}.

Assume now that \cC\ satisfies Inclusion and 2-Loop, and that 
\mbox{$A \subseteq B \subseteq \cC(A)$}. 
By Inclusion: \mbox{$A \subseteq B \subseteq \cC(B)$}. 
By 2-Loop, then, we have: \mbox{$\cC(A) = \cC(B)$}.
\QED
The finer study of C-logics relies, as for monotonic logics, 
on the notions of a consistent set and of a theory.
\begin{definition}
\label{def:con}
A set \mbox{$A \subseteq \cL$} is said to be {\em consistent}
iff \mbox{$\cC(A) \neq \cL$}.
A set $A$ for which \mbox{$\cC(A) = \cL$} is said to be {\em inconsistent}.
\end{definition}
The following follows from Idempotence.
\begin{lemma}
\label{le:C-con}
A set $A$ is consistent iff $\cC(A)$ is.
\end{lemma}
\begin{lemma}
\label{le:superinc}
If \mbox{$A \subseteq B$} and $A$ is inconsistent, so is $B$.
\end{lemma}
\proof
If \mbox{$\cC(A) = \cL$}, we have \mbox{$A \subseteq B \subseteq \cC(A)$} and,
by Cumulativity, \mbox{$\cC(B) = \cC(A) = \cL$}.
\QED
\begin{definition}
\label{def:maxcons}
A set \mbox{$A \subseteq \cL$} is said to be maximal consistent iff it is
consistent and any strict superset \mbox{$B \supset A$} is inconsistent.
\end{definition}
\begin{definition}
\label{def:theory}
A set \mbox{$T \subseteq \cL$} is said to be a {\em theory} iff 
\mbox{$\cC(T) = T$}.
\end{definition}
The following is obvious (by Inclusion).
\begin{lemma}
\label{le:uniq-inc}
There is only one inconsistent theory, namely \cL.
\end{lemma}
\begin{lemma}
\label{le:maxconsth}
Any maximal consistent set $A$ is a theory.
\end{lemma}
\proof
By Inclusion \mbox{$A \subseteq \cC(A)$}. By Lemma~\ref{le:C-con} the set
$\cC(A)$ is consistent and by maximality: \mbox{$A = \cC(A)$}.
\QED
\begin{notation}
Let us define \mbox{$\Cn : 2^{\cL} \longrightarrow 2^{\cL}$} by:
\[
\Cn(A) \ = \bigcap_{T \supseteq A , \ T {\rm \ a \ theory}} T.
\]
\end{notation}
The following follows from Lemma~\ref{le:uniq-inc}.
\begin{lemma}
\label{le:consT}
\[
\Cn(A) = \  \bigcap_{T \supseteq A , \ T {\rm \ a \ consistent \ theory}} T.
\]
\end{lemma}
\begin{lemma}
\label{le:supra}
\mbox{$A \subseteq \Cn(A) \subseteq \cC(A)$}.
\end{lemma}
\proof
By the definition of \Cn\ and the fact that $\cC(A)$ is a theory that includes 
$A$ (Idempotence and Inclusion).
\QED
\begin{lemma}
\label{le:CNinc}
A set \mbox{$A \subseteq \cL$} is inconsistent iff \mbox{$\Cn(A) = \cL$}.
\end{lemma}
\proof
If \mbox{$\Cn(A) = \cL$}, then, by Lemma~\ref{le:supra}, \mbox{$\cC(A) = \cL$}.
If $A$ is inconsistent, then, by Lemma~\ref{le:superinc}, there is no consistent
theory that includes $A$ and, by Lemma~\ref{le:consT}, \mbox{$\Cn(A) = \cL$}.
\QED
\begin{lemma}
\label{le:absorption}
\mbox{$\cC(A) = \Cn(\cC(A)) = \cC(\Cn(A))$}.
\end{lemma}
\proof
By Lemma~\ref{le:supra}, we have 
\mbox{$\cC(A) \subseteq \Cn(\cC(A)) \subseteq \cC(\cC(A))$}.
By Idempotence, then, the first equality is proved.
By Lemma~\ref{le:supra} and Cumulativity, we have \mbox{$\cC(A) = \cC(\Cn(A))$}.
\QED
\begin{corollary}
\label{co:T}
For any theory $T$, \mbox{$\Cn(T) = T$}.
\end{corollary}
\proof
\mbox{$\Cn(T) = \Cn(\cC(T)) = \cC(T) = T$}.
\QED
\begin{lemma}
\label{le:monCn}
The operation \Cn\ is monotonic, i.e., if \mbox{$A \subseteq B$}, 
then \mbox{$\Cn(A) \subseteq \Cn(B)$} and also idempotent, i.e., 
\mbox{$\Cn(\Cn(A)) = \Cn(A)$}.
\end{lemma}
\proof
Monotonicity follows immediately from the definition of \Cn.
For Idempotence, notice that, by Monotonicity and Corollary~\ref{co:T}, 
any theory $T$ that includes $A$ also includes $\Cn(A)$:
\mbox{$\Cn(A) \subseteq \Cn(T) = T$}.
\QED
\subsection{fC-models}
\label{subsec:fC-models}
Assume \cM\ is a set (of models), about which no assumption is made,
and \mbox{$\models \: \subseteq \cM \times \cL$} is a (satisfaction)
binary relation (nothing assumed either).
For any set \mbox{$A \subseteq \cL$}, we shall denote by 
\mbox{$\widehat{A}$} or by \mbox{${\rm Mod}(A)$}
the set of all models that satisfy all elements of
$A$:
\[
\widehat{A} = {\rm Mod}(A) = 
\{x \in \cM \mid x \models a , \: \forall a \in A \}.
\]
For typographical reasons we shall use both notations, sometimes even in
the same formula.
For any set of models \mbox{$X \subseteq \cM$}, we shall denote by 
\mbox{$\overline{X}$} the set of all formulas that are satisfied 
in all elements of $X$:
\[
\overline{X} = \{a \in \cL \mid x \models a , \forall x \in X \}.
\]
The following are easily proven, for any \mbox{$A , B \subseteq \cL$},
\mbox{$X , Y \subseteq \cM$}: they amount to the fact that the operations
\mbox{$X \mapsto \overline{X}$} and \mbox{$A \mapsto \widehat{A}$}
form a Galois connection.
\[
A \subseteq \overline{\widehat{A}} \ \ \ \ , \ \ \ \ X \subseteq \widehat{\overline{X}}
\]
\[
\widehat{A \cup B} = \widehat{A} \cap \widehat{B} \ \ \ , \ \ \ \overline{X \cup Y} = 
\overline{X} \cap \overline{Y}
\]
\[
A \subseteq B \Rightarrow \widehat{B} \subseteq \widehat{A} \ \ \ , \ \ \ 
X \subseteq Y \Rightarrow \overline{Y} \subseteq \overline{X}
\]
\[
A \subseteq B \Rightarrow \overline{\widehat{A}} \subseteq \overline{\widehat{B}} \ \ \ , \ \ \ 
X \subseteq Y \Rightarrow \widehat{\overline{X}} \subseteq \widehat{\overline{Y}}
\]
\[
\widehat{A} = \widehat{\overline{\widehat{A}}} \ \ \ , \ \ \ 
\overline{X} = \overline{\widehat{\overline{X}}}
\]
The last technical notion that will be needed is that of a definable set of
models. It will be used in the completeness proof below, but not in 
Definition~\ref{def:choicef}.
\begin{definition}
\label{def:definable}
A set $X$ of models is said to be definable iff one of the two
following equivalent conditions holds: 
\begin{enumerate}
\item
\mbox{$\exists A \subseteq \cL$} such that \mbox{$X = \widehat{A}$}, or
\item
\mbox{$X =  \widehat{\overline{X}}$}.
\end{enumerate}
The set of all definable subsets of $X$ will be denoted by $D_{X}$.
\end{definition}
The proof of the equivalence of the two propositions above is obvious.
\begin{lemma}
\label{le:definter}
If $X$ and $Y$ are definable sets of models, then their intersection 
\mbox{$X \cap Y$} is also definable.
\end{lemma}
\proof
By the remarks above: if \mbox{$X = \widehat{A}$} and
\mbox{$Y = \widehat{B}$}, \mbox{$X \cap Y =$}
\mbox{$\widehat{A} \cap \widehat{Y} =$}
\mbox{$\widehat{A \cup B} =$}.
\QED
\begin{definition}
\label{def:choicef}
A choice function on \cM\ is a function 
\mbox{$f : 2^{\cM} \rightarrow 2^{\cM}$}.
\end{definition} 
Note that $f$ is defined on arbitrary sets of models, not only on definable sets 
as in~\cite{L:LogicsandSemantics}.
In the same vein, we do not require here that the image by $f$ of a 
definable set be definable as was necessary in the corresponding soundness result 
of~\cite{L:LogicsandSemantics}.
\begin{definition}
\label{def:fC-models}
A triplet \mbox{$\langle \cM , \models , f \rangle$} is an fC-model (for language \cL) iff 
$\models$ is a binary relation on \mbox{$\cM \times \cL$} and $f$ is a choice function \cM\ that satisfies, for any sets $X$, $Y$:
\[
{\bf Contraction} \ \ \ f(X) \subseteq X
\]
and
\[
{\bf Local \ Cumulativity} \ \ \ \ f(X) \subseteq Y \subseteq X \Rightarrow
f(Y) = f(X).
\]
\end{definition}
\begin{definition}
\label{def:restricted-fC-models}
An fC-model is said to be a {\em restricted} fC-model iff its choice function
$f$ also satisfies, for any set $X$:
\[
{\bf Consistency} \ \ \ f(X) = \emptyset \Rightarrow
X  = \emptyset.
\]
\end{definition}
\subsection{Properties of fC-models}
This section makes clear the relation between fC-models and the models 
of~\cite{L:LogicsandSemantics}. It will not be used in the sequel.
There one considered choice functions satisfying
Contraction,
\[
{\bf Coherence} \ \ \ X \subseteq Y \Rightarrow X \cap f(Y) \subseteq f(X)
\]
and
\[
{\bf Local \ Monotonicity} \ \ \ \ f(X) \subseteq Y \subseteq X \Rightarrow
f(Y) \subseteq f(X).
\]
\begin{lemma}
\label{le:Coh}
Any function $f$ that satisfies Inclusion, Coherence 
and Local Monotonicity satisfies Local Cumulativity.
\end{lemma}
\proof
Assume \mbox{$f(X) \subseteq Y \subseteq X$}, we must show that
\mbox{$f(X) \subseteq f(Y)$} 
(the opposite inclusion is guaranteed by Local Monotonicity).
By Coherence: \mbox{$Y \cap f(X) \subseteq f(Y)$}.
\QED
\subsection{Soundness}
\begin{theorem}
\label{the:soundfC}
Let \mbox{$\langle \cM , \models , f \rangle$} be an fC-model and the 
operation \cC\ be such that: 
\begin{equation}
\label{eq:fdef}
\cC(A) = \overline{f(\widehat{A})}.
\end{equation}
Then \cC\ is a C-logics.
\end{theorem}
\proof
By Contraction \mbox{$f(\widehat{A}) \subseteq \widehat{A}$} and therefore
\mbox{$\overline{\widehat{A}} \subseteq \overline{f(\widehat{A})}$}.
But \mbox{$A \subseteq \overline{\widehat{A}}$}. 
We have proved Inclusion.

Assume now \mbox{$A \subseteq B \subseteq \cC(A)$}.
We have: 
\mbox{$\widehat{\cC(A)} \subseteq \widehat{B} \subseteq \widehat{A}$}.
But 
\mbox{$\widehat{\cC(A)} =$}
\mbox{$\widehat{\overline{f(\widehat{A})}}$}.
But \mbox{$f(\widehat{A}) \subseteq \widehat{\overline{f(\widehat{A})}}$}.
We have:
\mbox{$f(\widehat{A}) \subseteq \widehat{B} \subseteq \widehat{A}$}.
By Local Cumulativity, then:
\mbox{$f(\widehat{A}) = f(\widehat{B})$} and
\mbox{$\cC(A) = \cC(B)$}.
\QED
\subsection{Representation}
\begin{theorem}
\label{the:restr-rep}
If \cC\ is a C-logics, then there is a restricted fC-model 
\mbox{$\langle \cM , \models , f \rangle$} such that 
\mbox{$\cC(A) = \overline{f(\widehat{A})}$}.
\end{theorem}
Notice that, comparing to Theorem~\ref{the:soundfC} we are getting Consistency 
for free.
\proof
For \cM\ take all {\em consistent} theories of \cC.
Set \mbox{$T \models a$} iff \mbox{$a \in T$}.
It follows that, by Lemma~\ref{le:consT}, for any \mbox{$A \subseteq \cL$}, 
\[
\overline{\widehat{A}} = \Cn(A).
\]
We must now define a choice function $f$.
We shall do that in two stages. First, we shall define $f$ on definable sets
of models, then on arbitrary sets.
Suppose that $X$ is a definable subset of \cM\ and that
\mbox{$X = \widehat{A} = \widehat{B}$}.
Then, \mbox{$\Cn(A) = \Cn(B)$} and, by Lemma~\ref{le:absorption},
\mbox{$\cC(A) = \cC(B)$}.
The following definition of $f$, for any definable set of models, 
is therefore well-formed:
\mbox{$f(\widehat{A}) = \widehat{\cC(A)}$}.
It is worth noticing that for any $A$, $f(\widehat{A})$ is a definable set.
This is not in fact required by Theorem~\ref{the:soundfC}.
More stringent properties of $f$ on definable sets only, 
such as Coherence or the requirement that the
image by $f$ be a singleton, may be obtained but at the cost of 
definability-preservation.
We easily see that:
\mbox{$\overline{f(\widehat{A})} = $}
\mbox{$\overline{\widehat{\cC(A)}} = $}
\mbox{$\Cn(\cC(A) = $}
\mbox{$\cC(A)$}, by Lemma~\ref{le:absorption}.
Notice that, in the above, we use $f$ only on definable sets.
We shall now show that $f$ satisfies Contraction, Local Cumulativity and 
Consistency, when all sets considered are definable.
We shall leave for the end the definition of a proper extension of $f$
to arbitrary sets of models.
Let \mbox{$X = \widehat{A}$} be a definable subset of \cM.
By Inclusion, \mbox{$A \subseteq \cC(A) = \overline{f(\widehat{A})}$}.
Therefore \mbox{$\widehat{\overline{f(X)}} \subseteq X$}.
But \mbox{$f(X) \subseteq \widehat{\overline{f(X)}}$} as shown in 
Section~\ref{subsec:fC-models}. We have proved Contraction.

Let \mbox{$X = \widehat{A}$} and \mbox{$Y = \widehat{B}$} be 
definable subsets of \cL\ and assume:
\mbox{$f(X) \subseteq Y \subseteq X$}.
We have:
\mbox{$\Cn(A) \subseteq \Cn(B) \subseteq \cC(A)$}.
By Lemma~\ref{le:absorption} and Cumulativity, we have:
\mbox{$\cC(\Cn(A)) = \cC(\Cn(B))$} and, by Lemma~\ref{le:absorption},
\mbox{$\cC(A) = \cC(B)$}.
Therefore \mbox{$f(X) = \widehat{\cC(A)} = \widehat{\cC(B)} = f(Y)$}.
We have proved Local Cumulativity.

Let \mbox{$X = \widehat{A}$} be a definable subset of \cL\ such that 
\mbox{$f(X) = \widehat{\cC(A)} = \emptyset$}.
We have \mbox{$\Cn(\cC(A)) = \overline{\widehat{\cC(A)}} = \cL$}.
By Lemma~\ref{le:absorption}, then, $A$ is inconsistent.
By Lemma~\ref{le:superinc} there is no consistent theory that includes $A$ and
therefore \mbox{$\widehat{A} = X = \emptyset$}. We have proved Consistency.

We must now extend $f$ to arbitrary subsets of \cM\ in a way that enforces
Contraction, Local Cumulativity and Consistency.
Given an arbitrary subset \mbox{$X \subseteq \cM$}, we shall define $f'(X)$
by considering two cases.
\begin{itemize}
\item First, if there is some {\em definable} set $Y$ such that
\mbox{$f(Y) \subseteq X \subseteq Y$}, then we shall put \mbox{$f'(X) = f(Y)$}.
\item Secondly, if there is no such definable $Y$ we shall put
\mbox{$f'(X) = X$}.
\end{itemize}
We must first check that the first case above is a proper definition.
Suppose indeed that \mbox{$f(Y) \subseteq X \subseteq Y$} and
\mbox{$f(Z) \subseteq X \subseteq Z$} for definable sets $Y$ and $Z$.
Then, by Lemma~\ref{le:definter}, the set \mbox{$Y \cap Z$} is definable
and \mbox{$f(Y) \subseteq X \subseteq Y \cap Z \subseteq Y$}.
Therefore, by Cumulativity of $f$ (all sets considered are definable),
\mbox{$f(Y) = f(Y \cap Z)$}. 
Similarly, \mbox{$f(Z) = f(Y \cap Z)$} and \mbox{$f(Y) = f(Z)$}.

Let us notice now that, if $X$ is definable, then \mbox{$f'(X) = f(X)$},
since \mbox{$f(X) \subseteq X \subseteq X$} by Contraction.

Let us show, now, that \mbox{$f'(X) \subseteq X$}. 
In the first case: \mbox{$f'(X) = f(Y) \subseteq X$} and in the second case
\mbox{$f'(X) = X$}.

Suppose now that \mbox{$f'(Y) \subseteq X \subseteq Y$}, for arbitrary sets
$X$ and $Y$.
We shall consider the two different cases of the definition of $f'(Y)$.
If there is a definable $Z$ such that \mbox{$f(Z) \subseteq Y \subseteq Z$},
then we have \mbox{$f'(Y) = f(Z)$} and \mbox{$f(Z) \subseteq X \subseteq Z$}.
Therefore \mbox{$f'(X) = f(Z) = f'(Y)$}.
If there is no such $Z$, then \mbox{$f'(Y) = Y$} and \mbox{$X = Y$}.
Therefore \mbox{$f'(X) = f'(Y)$}.

For Consistency, assume \mbox{$f'(X) = \emptyset$}.
In the first case, there is a definable $Z$ such that 
\mbox{$f(Z) \subseteq X \subseteq Z$}. Then \mbox{$f(Z) = f'(X) = \emptyset$} and
\mbox{$Z = \emptyset$} and therefore \mbox{$X = \emptyset$}.
In the second case \mbox{$f'(X) = X = \emptyset$}.
\QED
\subsection{Connectives in C-logics}
\subsubsection{Conjunction and Negation}
We shall show that C-logics admit a classical conjunction 
and a classical negation. 
Let us assume now, for the remainder of this section, 
that the language \cL\ is closed under a binary connective 
written $\wedge$ and a unary connective written $\neg$.
\begin{theorem}
\label{the:andneg_sound}
If \mbox{$\langle \cM , \models , f \rangle$} is a restricted fC-model 
that behaves classically with respect to $\wedge$ and $\neg$,
i.e., for any \mbox{$m \in \cM$}, 
\begin{itemize}
\item \mbox{$m \models a \wedge b$} 
iff \mbox{$m \models a$} and 
\mbox{$m \models b$}
\item \mbox{$m \models \neg a$} iff \mbox{$m \not \models a$},
\end{itemize}
then the inference operation defined by the fC-model satisfies:
\begin{itemize}
\item {\bf $\wedge$-R} \mbox{$\cC(A , a \wedge b) = \cC(A , a , b)$}
\item {\bf $\neg$-R1} \mbox{$\cC(A , a , \neg a) = \cL$}
\item {\bf $\neg$-R2} if \mbox{$\cC(A , \neg a) = \cL$}, then \mbox{$a \in \cC(A)$}.
\end{itemize}
\end{theorem}
\proof
The first property follows from the fact that 
\mbox{${\rm Mod}(A \cup \{a \wedge b\}) =$}
\mbox{${\rm Mod}(A \cup \{a\} \cup \{b\})$}.
For the second property notice that
\mbox{${\rm Mod}(A \cup \{a\} \cup \{\neg a\}) = \emptyset$} implies, 
by Contraction, that 
\mbox{$f({\rm Mod}(A \cup \{a\} \cup \{\neg a\})) = \emptyset$}.
For the third property, since no element of \cM\ satisfies both $a$ and
$\neg a$, if \mbox{$\cC(A , \neg a) = \cL$}, 
there is no $m$ that satisfies \mbox{$\cC(A , \neg a)$} and 
\mbox{$f({\rm Mod}(A \cup \neg a)) = \emptyset$}.
Since the model is a restricted fC-model, 
\mbox{${\rm Mod}(A \cup \neg a) = \emptyset$}. 
Therefore every $m$ that satisfies $A$ also satisfies $a$ and
\mbox{$a \in \cC(A)$}. 
\QED
The reader should notice that it is claimed that, if
\mbox{$\cC(A , \neg a) = \cL$}, then any $m$ satisfying $A$ also satisfies
$a$, but it is {\em not} claimed that, under this hypothesis,
\mbox{$a \in \Cn(A)$}. Indeed \Cn\ is defined via the theories of \cC\ 
and the relation of those to the elements of \cM\ is not straightforward.
The reader should also note that a similar result (Equation 8.5) was obtained 
in~\cite{L:LogicsandSemantics} only assuming Coherence. Here Coherence 
is not required, Consistency is required in its place.
The following theorem shows the converse. It requires a compactness assumption.
We shall, then, assume that \cC\  satisfies the following:
\[
{\bf Weak \ Compactness} \ \ \ \ \cC(A) = \cL \: \Rightarrow \:
\exists {\rm \ a \ finite \ } B \subseteqf A \ {\rm such \ that} \ 
\cC(B) = \cL.
\]
\begin{theorem}
\label{the:andneg_comp}
If \cC\ satisfies Weak Compactness, Inclusion, Cumulativity, 
$\wedge$-R, $\neg$-R1 and $\neg$-R2,
then there is a restricted fC-model that behaves classically with respect to
$\wedge$ and $\neg$ such that \mbox{$\cC(A) = \overline{f(\widehat{A})}$}.
\end{theorem}
Before presenting a proof of Theorem~\ref{the:andneg_comp}, three lemmas are 
needed. 
A maximal consistent subset of \cL\ is a consistent subset that has no consistent superset.
\begin{lemma}
\label{le:wcomp}
Assume \cC\ satisfies Weak Compactness, $\neg-R1$ and $\neg-R2$. 
If \mbox{$a \not \in \cC(A)$}, there is a maximal consistent set 
\mbox{$B \supseteq A$} such that \mbox{$a \not \in B$}.
\end{lemma}
\proof
By $\neg-R2$, \mbox{$A \cup \{ \neg a \}$} is consistent.
By Weak Compactness (and Zorn's lemma), there is a maximal consistent set
$B$ that contains it. This $B$ does not contain $a$ by $\neg-R1$.
\QED
\begin{lemma}
\label{le:maxconsclass}
Assume \cC\ satisfies Inclusion, Cumulativity, $\wedge$-R, $\neg$-R1 and
$\neg$-R2.
If $A$ is a maximal consistent set, then
\begin{itemize}
\item \mbox{$a \wedge b \in A$} iff \mbox{$a \in A$} and \mbox{$b \in A$},
\item \mbox{$\neg a \in A$} iff \mbox{$a \not \in A$}.
\end{itemize}
\end{lemma}
\proof
By $\wedge$-R, \mbox{$a \wedge b \in \cC(A)$} iff \mbox{$a \in \cC(A)$} and
\mbox{$b \in \cC(A)$}, but,
by Lemma~\ref{le:maxconsth}, $A$ is a theory.
If \mbox{$\neg a \in A$}, then \mbox{$a \not \in A$} since $A$ is consistent,
by $\neg$-R1.
If \mbox{$\neg a \not \in A$}, then by the maximality of $A$, 
\mbox{$\cC(A , \neg a$} = \cL\ and, by $\neg$-R2, 
\mbox{$a \in \cC(A)$}, but $A$ is a theory.
\QED
\begin{lemma}
\label{le:Cnmax}
Assume \cC\ satisfies Weak Compactness, $\neg-R1$ and $\neg-R2$. 
Then
\[
\Cn(A) \ = \ \bigcap_{B \supseteq A , B {\rm \ maximal \ consistent}} B.
\]
\end{lemma}
\proof
The left-hand side is a subset of the right-hand side by 
Lemmas~\ref{le:consT} and~\ref{le:Cnmax}.
But if \mbox{$a \not \in \Cn(A)$}, then \mbox{$a \not \in \cC(A)$} and,
by Lemma~\ref{le:wcomp}, there is a maximal consistent $B$ that includes
$\cC(A)$ but does not contain $a$.
\QED
Let us now proceed to the proof of Theorem~\ref{the:andneg_comp}.
\proof
We modify the construction of Theorem~\ref{the:restr-rep}, by considering
not all consistent theories but only maximal consistent sets. 
Those maximal consistent sets are theories and behave classically for
$\wedge$ and $\neg$ by Lemma~\ref{le:maxconsclass}.
By Lemma~\ref{le:Cnmax}, for any \mbox{$A \subseteq \cL$}, 
\[
\overline{\widehat{A}} = \Cn(A).
\]
The remainder of the proof is unchanged.
\QED
We may now show that propositional nonmonotonic
logic is not weaker than (and therefore exactly the same as) monotonic
logic.
In the following theorem, we consider a propositional language 
in which negation and conjunction are considered basic and other connectives
are defined in the usual classical way. 
\begin{theorem}
\label{the:prop}
Let \cL\  be a propositional calculus (negation and conjunction basic, 
other connectives defined classically) and \mbox{$a , b \in \cL$}.
The following propositions are equivalent.
\begin{enumerate}
\item \label{logimpl}
$a$ logically implies $b$, i.e., \mbox{$a \models b$},
\item \label{mon}
for any operation \cC\ that satisfies Inclusion, Idempotence, Monotonicity,
Weak Compactness and the rules $\wedge$-R, $\neg$-R1 and
$\neg$-R2 above:
\mbox{$b \in \cC(a)$},
\item \label{si}
for any operation \cC\ that satisfies Inclusion, Cumulativity, 
Weak Compactness and the rules $\wedge$-R, $\neg$-R1 and
$\neg$-R2 above:
\mbox{$b \in \cC(a)$},
\item \label{forall}
for any such \cC\ and for any \mbox{$A \subseteq \cL$}:
\mbox{$b \in \cC(A , a)$},
\item \label{contra}
for any such \cC:
\mbox{$\cC(a , \neg b) = \cL$}.
\end{enumerate}
\end{theorem}
\proof
Property~\ref{contra} implies~\ref{forall},
since, by Cumulativity, \mbox{$\cC(a , \neg b) = \cL$}
implies \mbox{$\cC(A , a , \neg b) = \cL$}, and, by the
rule $\neg$-R2: \mbox{$b \in \cC(A , a)$}.
Property~\ref{forall} obviously implies~\ref{si}, that 
implies~\ref{mon} since Monotonicity and Idempotence imply Cumulativity.
It is easy to see that property~\ref{mon} implies~\ref{logimpl}.
Let $m$ be any propositional model that satisfies $a$.
Let \cC\  be defined by \mbox{$\cC(A) = \overline{\{m\}}$},
the set of formulas satisfied by $m$,
if \mbox{$m \in \widehat{A}$} and \mbox{$\cC(A) = \cL$}
otherwise.
By assumption, \mbox{$b \in \cC(a)$}.
But \mbox{$\cC(a) = \overline{\{m\}}$} since 
\mbox{$m \in \widehat{\{a\}}$}, therefore
\mbox{$m \models b$}.

The only non-trivial part of the proof is that~\ref{logimpl}
implies~\ref{contra}.
Assume \mbox{$a \models b$} and \cC\  satisfies
Inclusion, Cumulativity, Weak Compactness
and the rules $\wedge$-R, $\neg$-R1 and $\neg$-R2.
By Theorem~\ref{the:andneg_comp}, there is a set \cM, a satisfaction relation
$\models$ that behaves classically with respect to $\wedge$ and $\neg$ and
a definability-preserving choice function satisfying Contraction and
Local Cumulativity such that 
\mbox{$\cC(a , \neg b) =$}
\mbox{$\overline{f(\widehat{\{a\}} \cap \widehat{\{\neg b\}})}$}.
But, by assumption 
\mbox{$\widehat{\{a\}} \cap \widehat{\{\neg b\}} = \emptyset$}.
By Contraction, then
\mbox{$\cC(a , \neg b) =$}
\mbox{$\overline{\emptyset} =$} \cL.
\QED
Theorem~\ref{the:prop} shows that the proof theory of the 
semantically-classical conjunction and negation
in a nonmonotonic setting is the same as in a monotonic setting.
The following shows, that, in yet another sense, 
C-logics admit a proper
conjunction and a proper negation: one may conservatively extend any C-logics
on a set of atomic propositions 
to a language closed under conjunction and negation.
It is customary to consider Introduction-Elimination rules, such as
$\wedge$-R, $\neg$-R1 and $\neg$-R2
as definitions of the connectives.
Hacking~\cite[Section VII]{Hacking:What} discusses this idea
and proposes that, to be considered as bona fide definitions of the
connectives, the rules must be such that they
ensure that any legal logic on a small language may be
conservatively extended to a legal logic on the language extended
by closure under the connective.
\begin{theorem}
\label{the:conser}
Let $P$ be an arbitrary set of atomic propositions and \cC\ a C-logics over $P$.
Let \cL\  be the closure of $P$ under $\wedge$ and $\neg$.
Then, there exists a C-logics \cC' on \cL\ that satisfies
$\wedge$-R, $\neg$-R1 and $\neg$-R2, 
such that, for any \mbox{$A \subseteq P$},
\mbox{$\cC(A) = P \cap \cC'(A)$}.
\end{theorem}
\proof
By Theorem~\ref{the:restr-rep}, there is a restricted fC-model on $P$
\mbox{$\langle \cM , \models , f \rangle$} such that 
\mbox{$\cC(A) = \overline{f(\widehat{A})}$}.
Let us now extend $\models$ to \cL\, by \mbox{$m \models a \wedge b$}
iff \mbox{$m \models a$} and \mbox{$m \models b$} and 
\mbox{$m \models \neg a$} iff \mbox{$m \not \models a$}.
We claim that \mbox{$\langle \cM , \models , f \rangle$} is now a restricted
fC-model on \cL, whose satisfaction relation $\models$ behaves classically
for $\neg$ and $\wedge$. Indeed, the properties required from $f$ do not involve
the satisfaction relation at all, they deal with subsets of \cM\ exclusively.
Let us define, for any \mbox{$A \subseteq \cL$}, 
\mbox{$\cC'(A) = \overline{f(\widehat{A})}$}.
By Theorem~\ref{the:soundfC}, \cC'\ is a C-logics.
By Theorem~\ref{the:andneg_sound} it satisfies $\wedge$-R, $\neg$-R1 and $\neg$-R2. 
It is left to us to see that \mbox{$\cC(A) = P \cap \cC'(A)$},
for any \mbox{$A \subseteq P$}.
This follows straightforwardly from the fact that both $\cC(A)$
and $\cC'(A)$ are the sets of formulas (the former of $P$, the latter
of \cL) satisfied by all members of the set \mbox{$f(\widehat{A})$}. 
\QED
\subsubsection{Disjunction}
\label{sec:disj}
We have seen that any C-logics admits classical negation and conjunction.
The reader may think that this implies that it also admits a classical
disjunction defined as \mbox{$a \vee b = \neg ( \neg a \wedge \neg b )$}.
Indeed it is the case that, if we define disjunction in this way one of the
basic properties of disjunction is satisfied:
\[
{\bf \vee-R1} \ \ a \in \cC(A) \ \Rightarrow \ a \vee b \in \cC(A) 
{\rm \ and \ }
b \vee a \in \cC(A).
\]
But the other fundamental property of disjunction does not hold.
\[
{\bf \vee-R2} \ \ \cC(A , a) \cap \cC(A , b) \subseteq \cC(A , a \vee b).
\]
The following example shows that, in general, no proper disjunction can be
defined in C-logics.
\begin{example}
\label{ex:disj}
Consider the language \cL\ that contains four (atomic) propositions 
\mbox{$a, b , c , d$}.
Let \cM\ contain three elements: $m , n , p$.
Let $models$ be defined by: 
\mbox{$m \models a$}, \mbox{$m \models c$}, \mbox{$n \models a$},
\mbox{$n \models d$}, \mbox{$p \models b$} and \mbox{$p \models c$}.
The set \mbox{$\{m , n\}$} is definable (by $a$) and so is the set
\mbox{$\{m\}$} (by \mbox{$\{ a , c \}$}).
The function $f$ is the identity on all definable sets, except that 
\mbox{$f(\{m , n \}) = \{m\}$}.
The choice function preserves definability and satisfies Contraction and
Local Cumulativity. Indeed if \mbox{$f(X) \subseteq Y \subseteq X$} and
\mbox{$Y \neq X$}, we must have \mbox{$X = \{m , n\}$}
and \mbox{$Y = \{m\}$}. But in this case 
\mbox{$f(Y) = Y = f(X)$}. 
We have \mbox{$c \in \cC(a)$} and also \mbox{$c \in \cC(b)$}.
If there was a proper disjunction we should have also
\mbox{$c \in \cC(a \vee b)$}, and \mbox{$m \models a \vee b$},
\mbox{$n \models a \vee b$}
\mbox{$p \models a \vee b$}. 
Therefore \mbox{$\widehat{a \vee b} = \{m , n , p \}$} and
\mbox{$n in f(\{m , n , p \}$}.
But \mbox{$n \not \models c$}.
\end{example}
This example above has to be opposed to the results 
of~\cite{L:LogicsandSemantics} that show that if $f$ satisfies Coherence,
there is a satisfactory disjunction.
\subsection{Connection with previous work}
\begin{theorem}
\label{the:cumrelations_sound}
Let \cL\ be a propositional calculus and \cC\ an operation that satisfies
Weak-Compactness, Inclusion, Cumulativity, $\wedge$-R, $\neg$-R1 and $\neg$-R2.
Define a binary relation among propositions by:
\mbox{$a \NI b$} iff \mbox{$b \in \cC(a)$}.
Then, the relation \NI\ is a cumulative relation in the sense of~\cite{KLMAI:89}.
\end{theorem}
\proof
We shall show that \NI\ satisfies Left Logical Equivalence, Right Weakening,
Reflexivity, Cut and Cautious Monotonicity.
For Left-Logical-Equivalence, suppose \mbox{$\models a \leftrightarrow a'$}.
By Theorem~\ref{the:prop}, 
\mbox{$a' \in \cC(a)$} and, by Cumulativity, 
\mbox{$\cC(a) = \cC(a , a')$}.
But, similarly, exchanging $a$ and $a'$:
\mbox{$\cC(a') = \cC(a , a')$} and \mbox{$\cC(a) = \cC(a')$}.
For Right Weakening, by Theorem~\ref{the:prop} \mbox{$b \models b'$} implies
\mbox{$b' \in \cC(a , b)$}. If \mbox{$a \NIm b$}, by Cumulativity 
\mbox{$\cC(a) = \cC(a , b)$} and \mbox{$a \NIm b'$}.
Reflexivity follows from Inclusion.
Cut and Cautious Monotonicity together are equivalent to:
if \mbox{$a \NIm b$}, then \mbox{$a \wedge b \NIm c$} iff 
\mbox{$a \NIm c$}. Assume \mbox{$b \in \cC(a)$}, then, by Cumulativity,
\mbox{$\cC(a) = \cC(a , b)$}.
\QED
The converse also holds.
\begin{theorem}
\label{the:cumrelations_comp}
If \NI\ is a cumulative relation, then there is an operation \cC\ that satisfies
Weak-Compactness, Inclusion, Cumulativity, $\wedge$-R, $\neg$-R1 and $\neg$-R2
such that \mbox{$b \in \cC(a)$} iff \mbox{$a \NIm b$}.
\end{theorem}
\proof
The operation \cC\ may be defined in a way first proposed in the 
Theorem 14 of~\cite{FL:Studia}:
\mbox{$b \in \cC(A)$} iff there exists some formula $a$ such that
\mbox{$A \models a$} ($\models$ is logical implication of
propositional calculus) enjoying the following property:
for any $a'$ such that \mbox{$A \models a'$} and \mbox{$a' \models a$},
one has \mbox{$a' \NI b$}.
For Weak Compactness, assume \mbox{$\cC(A) = \cL$}. Then 
\mbox{${\bf false} \in \cC(A)$} and there is some $a$ such that
\mbox{$A \models a$} and \mbox{$a \NIm {\bf false}$}.
There is a finite subset $B$ of $A$ such that \mbox{$B \models a$}.
Let $b$ be the conjunction of all the propositions of $B$.
We have \mbox{$b \models a$} and therefore \mbox{$b \NIm a$}.
But \mbox{$a \NIm {\bf false}$} implies \mbox{$a \NIm b$}.
Therefore \mbox{$b \NIm {\bf false}$} and \mbox{$\cC(B) = \cC(b) = \cL$}.
The other properties claimed are not difficult to show.
\QED
\section{L-logics}
A sub-family of C-logics will be defined now. 
It corresponds to the cumulative with loop (CL) relations of~\cite{KLMAI:89}.
\begin{definition}
\label{def:L-logics}
The operation \cC\ is said to be an {\em L-logics} iff it satisfies the 
two following properties.
\[
{\bf Inclusion} \ \ \ \forall A \subseteq \cL\ , \ 
A \subseteq \cC(A) , 
\]
\[
{\bf Loop} \ \ \ \forall n \forall i = 0 , \ldots , n-1 {\rm \ modulo \ } n  
A_{i} \subseteq \cC(A_{i+1}) 
\Rightarrow \cC(A_{0}) = \cC(A_{1}).
\]
\end{definition}
The assumption of Loop is: \mbox{$A_{0} \subseteq \cC(A_{1})$},
\mbox{$A_{1} \subseteq \cC(A_{2})$}, $\ldots$ ,
\mbox{$A_{n-1} \subseteq \cC(A_{0})$}.
The conclusion could equivalently have been: \mbox{$\cC(A_{i}) = \cC(A_{j})$}
for any \mbox{$i , j =$} \mbox{$0 , \ldots , n-1$}.
Notice that for $n=2$, the condition Loop is the condition 2-Loop of 
Lemma~\ref{le:2-loop}. Therefore any L-logics is a C-logics.
The characteristic property of L-logics is embedded in the relation to be defined
now.
\begin{definition}
\label{def:<=}
Let $T$ and $S$ be theories. Let us define \mbox{$T \leq S$} iff
there exists a set \mbox{$A \subseteq S$} such that \mbox{$\cC(A) = T$}.
\end{definition}
The following holds without any assumption on \cC.
\begin{lemma}
\label{le:all<=}
The relation $\leq$ is reflexive.
If \mbox{$T , S$} are two theories such that \mbox{$T \subseteq S$}, then
\mbox{$T \leq S$}.
\end{lemma}
\proof
\mbox{$S \subseteq S$} and \mbox{$\cC(S) = S$} imply \mbox{$S \leq S$}.
\mbox{$T \subseteq S$} and \mbox{$\cC(T) = T$} imply \mbox{$T \leq S$}.
\QED
The next lemma holds only for L-logics.
Notice that, even for L-logics, the relation $\leq$ is not transitive in general.
\begin{lemma}
\label{le:L<=}
If \cC\ is an L-logics, and
\mbox{$T_{0} \leq T_{1}$}, $\dots$, \mbox{$T_{n-1} \leq T_{0}$}, then 
\mbox{$T_{0} = T_{1} =$} \mbox{$\ldots = T_{n-1}$}.
\end{lemma}
\proof
Assume \mbox{$T_{0} \leq T_{1}$}, $\dots$, \mbox{$T_{n-1} \leq T_{0}$}.
There are \mbox{$A_{i} \subseteq T_{i+1}$} such that
\mbox{$\cC(A_{i}) = T_{i}$}.
Therefore \mbox{$A_{i-1} \subseteq \cC(A_{i})$} and by Loop
\mbox{$\cC(A_{i}) = \cC(A_{j})$}.
\QED
In particular, the relation $\leq$ is antisymmetric for L-logics 
(in fact for C-logics).
\begin{definition}
\label{def:<}
Let $T$ and $S$ be theories. Let us define \mbox{$T < S$} iff
\mbox{$T \leq S$} and \mbox{$S \not \leq T$}, or equivalently (for C-logics)
\mbox{$T \leq S$} and \mbox{$T \neq S$}.
Let $<^{+}$ be the transitive closure of $<$.
\end{definition}
\begin{lemma}
\label{le:<+}
If \cC\ is an L-logics, then the relation $<^{+}$ is irreflexive and therefore
a strict partial order.
\end{lemma}
\proof
By Lemma~\ref{le:L<=}.
\QED
\section{Quantum Consequence Operations}
Birkhoff and von Neumann~\cite{BirkvonNeu:36} framed Quantum Logics in 
Hilbert style, i.e., as a set of valid propositions in propositional
calculus. Engesser and Gabbay~\cite{EngGabbay:Quantum} proposed to view 
Quantum Logics in a different light: as a consequence relation describing 
what can be deduced from what. 
They assume a language closed under the propositional connectives, but
their definition makes perfect sense and is very rich even on a language
that contains only atomic propositions.
This is, in this paper's view, a major step taken by Engesser and Gabbay
since Birkhoff and von Neumann's framework does not allow 
any interesting consideration in the absence of connectives.
The setting proposed by Engesser and Gabbay allows us to discuss first the
nature of Quantum Deduction without any need to posit connectives, and then
to consider the proof-theoretic and semantics properties of connectives
one at a time.

Assume a Hilbert space \cH\ and an element \mbox{$h \in \cH$} are given.
Assume also a non-empty set (language) \cL\ of closed subspaces of \cH\ 
is given.
The elements of \cL, the atomic propositions are, thus, 
closed subspaces of \cH.
For every proposition \mbox{$a \in \cL$}, we shall denote by \mbox{$a_{p}$} 
the projection on the subspace $a$: for every \mbox{$x \in \cH$}, 
\mbox{$a_{p}(x)$} is the element of $a$ closest to $x$.
For every set of propositions: \mbox{$A \subseteq \cL$}, 
\mbox{$A^{*} \eqdef \bigcap_{a \in A} a$} and \mbox{$A^{*}_{p}$} will
denote the projection on $A^{*}$, i.e.,
on the intersection of all the elements of $A$.
\begin{definition}[Engesser-Gabbay]
Let \mbox{$\cC : 2^{\cL} \longrightarrow 2^{\cL}$} be defined by:
\begin{equation}
b \in \cC(A) {\rm \ iff \ } A^{*}_{p}(h) \in b.
\end{equation}
\end{definition}
\begin{theorem}
\label{the:QL}
The operation \cC\ defined above is an L-logics.
\end{theorem}
Engesser and Gabbay essentially noticed already that \cC\ is a C-logics.
We need a lemma.
\begin{lemma}
\label{le:BCA}
If \mbox{$B \subseteq \cC(A)$}, then 
\mbox{$A^{*}_{p}(h) = (A^{*} \cap B^{*})_{p}(h)$} and
\mbox{$d(h , A^{*}) \geq d(h , B^{*})$}.
\end{lemma}
\proof
For any \mbox{$b \in B$}, \mbox{$A^{*}_{p}(h) \in b$}.
Therefore \mbox{$A^{*}_{p}(h) \in B^{*}$}.
\QED
Let us now prove Theorem~\ref{the:QL}
\proof
Indeed, \mbox{$A^{*}_{p}h \in A^{*}$} and therefore, for any
\mbox{$a \in A$}, \mbox{$A_{p}h \in a$}, and we have shown Inclusion.

Assume \mbox{$A \subseteq B \subseteq \cC(A)$}.
By Lemma~\ref{le:BCA}, we have 
\mbox{$A^{*}_{p}(h) = (A^{*} \cap B^{*})_{p}(h)$}, but 
\mbox{$B^{*} \subseteq A^{*}$} and
\mbox{$A^{*}_{p}(h) = B^{*}_{p}(h)$}. Therefore \mbox{$\cC(A) = \cC(B)$}
and we have shown Cumulativity.

\[
A_{1} \subseteq \cC(A_{0}) , A_{2} \subseteq \cC(A_{1}) , \ldots , 
A_{0} \subseteq \cC(A_{n}) \Rightarrow
\cC(A_{0}) = \cC(A_{1})
\]

For Loop, assume \mbox{$A_{i} \subseteq \cC(A_{i+1})$}, for 
\mbox{$i = 0 , \ldots , n - 1$} (mod $n$).
By Lemma~\ref{le:BCA},
\mbox{$d(h , A_{i+1}) \geq d(h , A_{i})$} 
and therefore all those distances are
equal: \mbox{$d(h , A_{0}) = d(h , A_{1})$}, 
\mbox{${A_{0}}^{*}_{p}(h) =  {A_{1}}^{*}_{p}(h)$} and 
\mbox{$\cC(A_{0}) = \cC(A_{1})$}.
\QED
\section{Open Question}
Do the four properties above characterize those consequence operations 
presentable by Hilbert spaces? 
Or are there other properties shared by those operations presentable by Hilbert
spaces that do not follow from the above?
\section{Connectives}
\subsection{Conjunction}
Conjunction is unproblematic. Even infinite conjunctions are easily defined.
If $A$ is a set of propositions, and each \mbox{$a \in A$} is associated with
some closed subspace $a^{*}$, we may associate the proposition 
\mbox{$\bigwedge_{a \in A} a$} with the closed subspace
\mbox{$\bigcap_{a \in A} a^{*}$}, i.e., $A^{*}$ and the rule $\wedge$-R
is validated: \mbox{$\cC(A , B) = \cC(\bigwedge a , B)$}.
\subsection{Negation}
The situation for negation is most intriguing.
By Theorem~\ref{the:QL} any operation \cC\ presented as a Quantum Logic is
an L-logics, therefore a C-logics.
Theorem~\ref{the:conser} shows that C-logics admit
a negation satisfying $\neg$-R1 and $\neg$-R2. 
We therefore expect Quantum Logics to admit such a negation. 
But the treatment of negation proposed
by Birkhoff and von Neumann and later used by Engesser and Gabbay does
not do the job in the following sense.
Suppose we define \mbox{$(\neg a)^{*} = (a^{*})^{\perp}$} where $\perp$ denotes
the orthogonal complement. It is easy to see that $\neg$-R1 is satisfied
since the intersection of a subspace and its orthogonal complement is $\{0\}$,
but $\neg$-R2 is not satisfied. 
Consider for example three generic (not parallel and not orthogonal) 
one-dimensional subspaces (lines through the origin) $a$, $b$ and $c$ 
in the real plane.
Let $h$ be any non-zero vector of $c$.
The intersection of $a$ and $b^{\perp}$ is $\{0\}$ and therefore
\mbox{$\cC(a , \neg b) = \cL$}.
But \mbox{$b \not \in \cC(a)$} since the projection of $h$ on $a$ 
is not in $b$.
This failure of $\neg$-R2, which is the principle of proof by contradiction,
was in fact already noted or guessed by Birkhoff and von Neumann.
In section 17, p. 837, they compare Quantum Logics with other non-classical
logics introduced on introspective or philosophical grounds, 
such as intuistionistic logics. They note that even though ``logicians
have usually assumed that properties of negation were the ones  least
able to withstand a critical analysis, the study of (quantum) mechanics
points to the distributive identities as the weakest link in the algebra
of logic.'' And they conclude: ``our conclusion agrees perhaps more with those
critiques of logic, which find most objectionable the assumption that 
to deduce an absurdity from the conjunction of $a$ and not $b$, justifies one 
in inferring that $a$ implies $b$''.
This paper's conclusions agree only in part, and will be presented below. 
\subsubsection{Disjunction}
A proper disjunction should satisfy $\vee$-R1 and $\vee-R2$ defined in 
Section~\ref{sec:disj}. 
We have seen that C-logics do not always support such a disjunction.
It is left to be seen whether Quantum Logics support such a disjunction.
In any C-logics that satisfies $\wedge$-R, $\vee$-R1 and $\vee$-R2, 
the distributive equality holds, in the sense that
\mbox{$\cC(A , a \wedge ( b \vee c )) = $}
\mbox{$\cC(A , (a \wedge b) \vee (a \wedge c))$}.
The only Quantum Logics that admit a proper disjunction are therefore
those Quantum Logics that support the distributive law. 
This is a very limited family. 
\section{Conclusions and future work}
Quantum Logics are nonmonotonic logics as noticed by Engesser and Gabbay,
they are also very respectable nonmonotonic logics since they are L-logics.
It is indeed surprising that Quantum Logics come to satisfy formal properties
designed with a completely different intention: to describe properties
``introduced on introspective grounds'' and intended to describe disciplined
``jumping to conclusions''. 
Intersection of closed subspaces provides a perfect semantics for conjunction.
Orthogonal complement does not provide a suitable semantics for negation,
but there is probably a respectable negation. 
It seems doubtful that one could find a suitable corresponding  operation 
among closed subspaces of Hilbert spaces that would enable us to associate
a closed subspace to the negation of a closed subspace.
This probably means that one cannot assume that the negation of an observable
is an observable. 
But must we insist that the negation of an observable be observable? 
Couldn't negation mean something about what we know and not about the world?
Disjunction is probably incompatible with Quantum Logics altogether.
\section{Acknowledgments}
I am most grateful to Kurt Engesser for helping me through the geometry
of Hilbert spaces. Some of the examples used in the paper are his.
\bibliographystyle{plain}

\end{document}